# Intrinsic Motivations and Open-ended Learning

**Gianluca Baldassarre**

Laboratory of Computational Embodied Neuroscience, Istituto di Scienze e Tecnologie della Cognizione, Consiglio Nazionale delle Ricerche (LOCEN-ISTC-CNR),

Via San Martino della Battaglia 44, I-00185 Roma, Italy

gianluca.baldassarre@istc.cnr.it

**Abstract**

There is a growing interest and literature on intrinsic motivations and open-ended learning in both cognitive robotics and machine learning on one side, and in psychology and neuroscience on the other. This paper aims to review some relevant contributions from the two literature threads and to draw links between them. To this purpose, the paper starts by defining intrinsic motivations and by presenting a computationally-driven theoretical taxonomy of their different types. Then it presents relevant contributions from the psychological and neuroscientific literature related to intrinsic motivations, interpreting them based on the grid, and elucidates the mechanisms and functions they play in animals and humans. Endowed with such concepts and their biological underpinnings, the paper next presents a selection of models from cognitive robotics and machine learning that computationally operationalise the concepts of intrinsic motivations and links them to biology concepts. The contribution finally presents some of the open challenges of the field from both the psychological/neuroscientific and computational perspectives.

# Introduction

There is wide and growing literature on *intrinsic motivations* (IMs) from both the biological perspective, involving in particular psychology and neuroscience, and the computational perspective, involving cognitive robotics and machine learning. The objective of this paper is to review some contributions from these two threads of the literature and to draw links between them. The research carried out within the two perspectives is often carried out in parallel, with limited cross talk between the two. A more intense dialogue between the two could lead to benefit both. On one side, the biological research might benefit of operational concepts and models of IMs from the computational perspective to produce quantitative detailed predictions to test in experiments. On the other side, computational research might benefit of high-level insights on the general principles behind natural IM systems that might suggest new ways to structure the architecture of artificial IM systems, and that might suggest new classes of algorithms accomplishing useful computational functions within open-ended learning. *Open-ended learning* refers to robots and agents that, similarly to the early development of humans (Weng et al. 2001, Lungarella et al. 2003), undergo prolonged periods of learning where they autonomously acquire knowledge and skills that might be useful to later solve extrinsic tasks given by the user (Baldassarre and Mirolli 2013; Doncieux et al. 2018).

The paper is organised as follows. First it gives a computationally-driven theoretical grid of the concepts, in particular by defining IMs by contrasting them with *extrinsic motivations* (EMs), and then by giving a classification of different IMs based on the mechanisms they rely on and the typical cognitive functions they might serve, in particular by referring to three main classes of IMs called here *epistemic intrinsic motivations* (eIM). This grid is then used to draw the links between the biological and computational perspectives. Second, it presents some relevant psychological and neuroscientific literature to exemplify possible behavioural functions of the different classes of eIMs, and the possible neural mechanisms underlying them. Third, it presents a selection of contributions from cognitive robotics and machine learning that exploit IMs, both eIMs and other IMs, to face computationally relevant problems; this also allows the computational operationalisation of the biological concepts on IMs thus drawing the seeked links between the biological and computational perspectives on them. The work concludes by presenting some of the open challenges of the research on IMs within the two perspectives.

# A conceptual grid: extrinsic and intrinsic motivations mechanisms and functions, and classes of (epistemic) intrinsic motivations

The concept of IM has been proposed and developed within the psychological literature to overcome the difficulties of the behaviourist theory on learning and drives (e.g., Skinner, 1938; Hull, 1943), in particular to explain why animals spontaneously engage in puzzles (Harlow, 1950) or can be instrumentally conditioned to produce particular responses on the basis of apparently neutral stimuli (e.g., a sudden light onset; Kish, 1955) as it happens with "standard" primary rewards (e.g., food). Subsequent proposals highlighted how the properties of certain *stimuli* can indeed trigger animal exploration and guide their learning processes, for example when the stimuli are complex, unexpected, or in general ``surprising'' (Berlyne, 1966). Another important thread of psychological research highlighted the importance that *action* plays in IMs, for example in relation to the fact that an agent manages to affect the environment with its behaviour (*effectance*; White, 1959), or can autonomously set own *goals* and master their achievement (Ryan, 2000).

Within the computational sciences, Schmidhuber (1991) firstly presented a computational operationalisation of some IM mechanisms (in particular *prediction-based IMs*, see below), and Barto et al. (2004), settled the fundamental link between IMs (in particular *competence-based IMs*, see below) and *reinforcement learning* (RL) methods (Sutton and Barto 2018). These ideas were first developed within the Developmental Robotics scientific community (with works in the journal "IEEE TAMD" and the conferences "ICDL" and "EpiRob"; Zlatev and Balkenius 2001, Lungarella et al. 2003, Oudeyer et al. 2007b, Schembri et al. 2007, Doya and Taniguchi 2019), and lately were developed within the autonomous/cognitive robotics and machine learning community (e.g., Bellemare, 2016; Nair et al. 2018), in particular driven by the success of deep neural networks and RL (Goodfellow et al. 2017; Sutton and Barto 2018).

Now the concepts used here are presented more in detail. These concepts will support the discussion of representative contributions from the biological and computational perspectives and the illustration of the links between them. The concepts of this grid often refer to the two key points of view from which one can look at cognition elements and computational models, namely (Tinbergen, 1963; Marr and Poggio, 1976): (b) the *computational functions* they serve, i.e. the *problems* they solve: the functions indicate the possible "uses" for which they might be employed within an overall cognitive/robotic system; (a) the *mechanisms,* or *algorithms,* used to accomplish those functions: the mechanisms refer to brain operations or the algorithms that are used as a means to actually accomplish the functions. Some specifications are due in relation to the use of the terms "functions" and "mechanisms" done here. First, for animals "function" is referred to *adaptive function*, i.e. the utility of certain elements of intelligence, such as an IM, for the animal *biological fitness*. For robots, "function" is referred to the utility of a certain element of the robot's intelligence for the robot's *user*. Second, as the functions of a computer program, "functions" can be organised at multiple hierarchical levels: from the highest level just mentioned ("biological fitness"; "utility for the user"), to lower levels, for example "producing movements to change the world as desired", "memorising precepts", "anticipating future percepts", etc.; each of these functions can be further decomposed into even lower-level functions, for example "changing the world" might require to "recognise the current state of the world", "anticipate a desired state of the world (goal)", and "performing an action to lead the world from the current to the desired state". Thus, a function can be seen as realised through a mechanism, but this mechanism in turn can be seen as a function to be realised, and as such it can be solved by lower-level mechanisms: this downward decomposition can continue until mechanisms that are (arbitrarily) considered as primitive for a given analysis. For this reason, here the term "mechanism" is used to mean the lowest and most specific level of the hierarchy of the considered functions, considerable as primitive for the given purposes.

## Extrinsic and intrinsic motivations

What are *motivations*? Motivations are an element of intelligence having at least three important functions (for organisms, cf. Panksepp 1998): (a) *selection*: drive the system to select a behaviour, among alternative available ones, to attend the most important current needs/goals; (b) *energy*: establish the amount of energy invested in executing the selected behaviour; (c) *learning*: generate learning signals to change behaviour. This paper considers in particular the first and third function of motivation. For example, we will see how IMs can drive an agent to move to some areas of the environment in navigation tasks (behaviour selection), or can produce the reward signals for reinforcement learning processes (production of learning signals).

What are *intrinsic motivations*? When initially studied in psychology, IMs were defined as motivations driving the performance of behaviour "for its own sake", i.e. without any direct apparent purpose (Berlyne 1966). Although useful to guide intuition, this definition does not clarify the functions, nor the mechanisms, of IMs. A more operational definition proposed here is that *intrinsic motivations are processes that can drive the acquisition of knowledge and skills in the absence of extrinsic motivations*. IMs are hence best understood by contrasting them to *extrinsic motivations* (EMs). Table 1 highlights the main differences between EM and a very important subset of IMs that are called here *epistemic intrinsic motivations (eMSs).* In Baldassarre (2011) EIM were considered to be IMs *tout court*, but here it is recognised that they do not cover the full spectrum of IMs because, as we shall see, there are some IMs that are not eIMs. In the table, EMs are however contrasted to eIMs because these form the bulk of IMs and because for their distinctive features they can help to clarify the overall nature of the whole set of IMs with respect to EMs.

Regarding functions, EMs have the overall function of driving behaviour and learning to the *acquisition of material resources*. For example, the EM of "hunger" drives behaviour to look for and ingest food, and when this happens the behaviour leading to it is strengthened. Instead, IMs have the overall function of driving behaviour and learning towards the *acquisition of knowledge and skills* (knowledge also encompasses skills, but these are mentioned explicitly to emphasize the aspects of knowledge more directly linked to action). For example, an IM related to novelty seeking could drive an agent to explore a novel object. In this respect, IMs have an *epistemic function*. This function is shared by *all* IMs, not only by eIMs, as *all* IMs support the acquisition of knowledge and skills: in other words, all IMs have an epistemic function. In this respect, the term "epistemic motivations" might have been used in place of the term IMs, misleading as "intrinsic" suggests "internal" or at best -and stretching it- to "not directed to external material resources", but the term IMs is kept for its tradition. Moreover, the term eIMs is handy to refer to IMs that, as shown below, are also based on an "epistemic mechanism" whereas non-epistemic IMs do not do so. In this respect, eIMs are the most prototypical IMs and a term referring only to them is useful.

**Table 1**: Main features of extrinsic and (epistemic) intrinsic motivations.

|  | **Extrinsic motivations (EMs)** | **(Epistemic) Intrinsic motivations (eIMs)** |
|---|---|---|
| **Function** | Organisms: acquisition of *material resources*. Robots: accomplishment of *user's goals*. | Acquisition of *knowledge and skills*. |
| **Mechanism** | Organisms: measure the acquisition of *material resources* by getting *information* on their levels/changes *from body* and *resource monitoring*. Robots: measure the level/change of accomplishment of the user's goals. | Measure the acquisition of knowledge and skills by getting *information* on their levels/changes in *other parts of the brain (organisms)* or *controller (robots)*. |
| **Time of contribution to the function** | *Immediately*: when the material resource are achieved and used (organisms); when the user's goals are accomplished (robots). | *Later*: when the acquired knowledge and skills are used to acquire resources (organisms) or to accomplish the user's goals (robots). |
| **"Time signature" of the motivation** | They tend to *go away* when the related resources are acquired, and to *come back* when they lack. | They tend to *go away for good* when the related pieces of knowledge/skills are acquired. |

In terms of *mechanisms*, in animals EMs are based on measures of the acquisition of material resources by getting information on their levels/changes *in the body* or *in the environment*. For example, hunger (a drive guiding the selection of behaviours related to food seeking) might be triggered when the blood glucose levels are low; and a reward signal (a learning signal) might be produced when food is ingested. Alternatively, an EM might be related to monitoring the presence/absence of resources externally to the body, for example the presence of a mating companion or the smell of prey in the environment (Baldassarre 2011). In robots, EMs are based on the measure of the accomplishment of the user's goals. Here the terms "extrinsic tasks/goals" will thus be referred to tasks/goals involving the accomplishment of material resources or user's goals. Incidentally, notice how EMs are a direct derivation of evolution: in animals, the acquisition of material resources is a means to increase biological fitness (number of fertile offspring), and more specifically the means for it, i.e. survival and reproduction; similarly in robots, the successful accomplishment of the user's goals produces a higher chance that the specific features of the robot controller and physical structure are "reproduced", as they are or in variants, in future robots.

Instead, eIMs rely on mechanisms that measure knowledge and skills by getting information on their levels/changes *in other parts of the brain* (for organisms) or *in controller* (for robots). Importantly, note how this implies that an eIM involves the presence of at least three structures and functions inside the brain/controller: (a) a "*source-component*" that acquires knowledge; (b) an *IM-mechanism* that receives in input information on the level or change of the knowledge of source-component; (c) a "*target-component*" that receives the output of the IM-mechanism and uses it to select behaviour/energise behaviour/drive learning processes. The core of the whole eIM process is the IM mechanism that measures the level or change of knowledge of the source-component.

The specification above is very important as conceptually eIMs involve the learning processes and knowledge to *two* different cognitive/computational components that might be very different in terms of mechanisms and functions they play within the overall system, and this might make difficult to recognise them in organisms or to implement them in robots. In some cases (Figure 1), the source-component and the target-component are the same data structure, in the sense that the IM-mechanism detects the knowledge level/change in a component with the function of affecting the learning of the same component (possibly with the mediation of other components). For example, the hippocampus might detect the novelty of an item and this might favour the memorisation of the item by the hippocampus itself (Lisman and Grace 2005). As another example, from robotics (Santucci et al. 2016), the selection of the skill to train among many skills to be learned might be based on the competence improvement of the skills themselves. In other cases (Figure 1), the sources-component and target components are distinct. For example, the hippocampus (or a component of a robot) might detect the novelty of some objects and this might drive a motor component to explore them with the function of improving its motor ability to manipulate them.

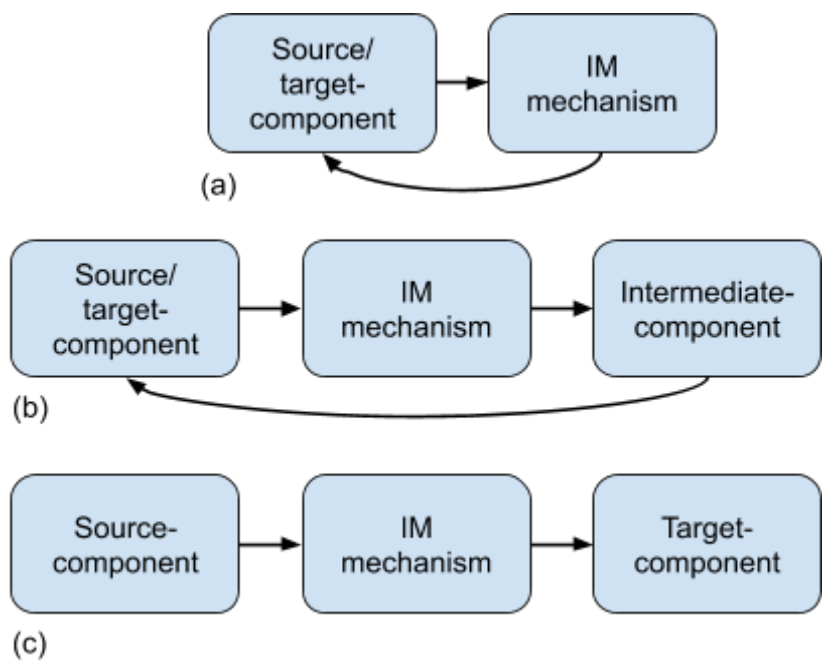

**Figure 1**: The key components of eIMs. (a) Case where the source-component and target-component are the same structure. (b) Case where the source-component and the target-component are the same structure but the retro-action is mediated by an intermediate component; (c) Case where the source-component and the target-component are different structures.

IMs that are not eIMs differ from the latter as they do not use a learning source-component as the origin of the motivation, but rather another mechanism: these will be called *other IMs* (oIMs) to distinguish them from eIMs. Sometimes such "other" mechanism mimics the acquisition of knowledge by a possible source-component, but the latter is not actually present. For example, count-based novelty mechanisms (Bellemare et al. 2016) perform novelty detection on the basis of the frequency with which states are encountered rather than on the basis of how well they are memorized (although it is true that they are still present/absent in the counter memory). In other cases, other mechanisms are used that have the function of acquiring knowledge and skills but do not seem related to knowledge acquisition as a trigger. For example, as discussed in the following sections, the principle of *empowerment* (Klyubin et al. 2005), or the concept of *bottlenecks* (McGovern Barto 2001), can support the acquisition of skills but they do not measure the knowledge of a source-component but rather some properties of the environment or of the agent's actions.

A critical difference between EMs and all IMs is the time *when* they express their function (utility). EMs tend to express their function at a time very proximal to when they are triggered. This because they lead to accomplish and consume material resources (organisms), or to accomplish the user's goals (robots), and when this happens they manifest their utility. Instead, IMs lead to accomplish knowledge and skills that are useful only later with respect to the time when they operate, in particular when such knowledge and skills are used to accomplish material resources or solve user's goals. This point is particularly important as it makes difficult to actually measure the effectiveness of a given IM. A possible way to measure such effectiveness is to divide the life of the agent into two phases (Schembri et al. 2007; Baldassarre et al. 2019): (a) *Intrinsic motivation phase*: the agent uses IMs to acquire knowledge and skills without a direct utility; (b) *Extrinsic motivation phase*: the agent uses the knowledge and skills acquired in the intrinsic phase to solve extrinsic problems. These two phases resemble the two main phases of human life involving a first infancy/childhood phase, mainly guided by IMs, and an adulthood phase, mainly guided by EMs (Schembri et al. 2007). The idea of the two phases was also set at the core of the *REAL competition*[1] (Robot open-Ended Autonomous Learning) proposed to create a benchmark for open-ended learning. In this competition, during a first intrinsic phase a simulated camera-arm-gripper robot can freely interact with some objects to autonomously acquire knowledge and skills without being given any goal or reward; then in a second extrinsic phase the quality of such knowledge and skills is measured by asking the robot to solve some sampled extrinsic tasks involving the re-creation of some sampled object configurations. Two caveats are due on this issue. Often in organisms, but also robots, IM and EM operate in parallel, for example a robot might aim to learn how to manipulate an object while accomplishing a user's task. This requires suitable *arbitration mechanisms* to mediate between IMs and EMs. Second, IMs and EMs mechanisms and functions might be mixed: for example the "source-component + IM-mechanism" might support a "target-component" pursuing an extrinsic goal. For example, next sections show a common use of novelty-based IMs to improve exploration in the accomplishment of extrinsic RL tasks involving "sparse rewards".

EMs and eIMs (and sometimes also oIMs) also have a typical "temporal signature" (Baldassarre, 2011). In particular, EMs tend to go away when the resources they are directed to have been obtained, and to come back when the resources are consumed/lost. For example, hunger and the reward of food ingestion go away after a sufficient amount of food is ingested, and to come back when, say, the blood glucose level are low again. Instead, eIMs triggered by the acquisition of a particular piece of information stored in the source-component tend to go away for good when such piece of information is acquired (unless the information is forgotten). From a biological perspective, this helps to recognise if a motivation is an (e)IM or an EM; from a computational perspective it is relevant as it possibly makes problems non-stationary, hence more challenging.

---

[1] https://www.aicrowd.com/challenges/neurips-2019-robot-open-ended-autonomous-learning

Having drawn the distinction between IMs and EMs, it is also useful to consider some aspects that are not relevant for such distinction. First, IMs should not be confused with motivations related to *secondary rewards* studied within psychology (Lieberman 1993). These are initially-neutral stimuli that acquire reinforcing properties based on a repeated association with *primary rewards* (e.g., a bell ring that anticipates several times the delivery of a food to a dog becomes motivating for it). In this respect, both EM and IM generates primary rewards, i.e. rewards that are not related to the association to other rewards but are generated by a primitive mechanisms originating "value" for the organism/robot (i.e., they are linked to the acquisition of a material resources, or to the solution of the user's goal, or to the acquisition of a piece of information). Another irrelevant distinction is between "internal and external" elements: both EMs and IMs involve stimuli that are both *external* to the agent (e.g., a food/body change, or a novel object) and mechanisms that are *internal* to it and that generate the motivation (Barto et al. 2004), as shown at length below. It is true, however, that IMs are more "inward looking" as they measure knowledge inside the brain/controller whereas EMs measure resources in the body or in the environment.

## Three classes of eIMs

The computational literature has given a strong contribution to distinguish between different classes of IM mechanisms. These classes in particular involve eIMs, and often are not applicable to oIMs: here the classification refers to IMs *tout-court* to stay with the standard nomenclature. A first contribution (Oudeyer and Kaplan 2007) distinguishes between *knowledge-based IMs*, related to the acquisition of information on the world, and *competence-based IMs (CB-IM)*, related to the acquisition of the capacity to act effectively. Another contribution (Barto et al. 2013) highlights the need to differentiate between two types of knowledge-based IMs, namely novelty-based IMs (NB-IM) and prediction-based IMs (PB-IM), often confused within the computational and biological literature. Now the main features of the three classes of IMs, summarised in Table 2, are considered more in detail. The classes are based on the function played by the source-component. For each class, there exist many sub-classes depending on the functions and mechanisms of the target-components. The IM-mechanism always measures the level or change of the knowledge of the source-component.

NB-IMs are based on a *memory* source-component that encodes patterns, e.g. percepts, with the function of storing and possibly re-coding them in more useful formats, e.g. to compress information or to facilitate down-stream processes. The IM-mechanism of NB-IMs performs a measurement of the source-component based on a one-step process: this checks the level of *novelty/familiarity* of a target pattern, e.g. a percept from the world. Another possibility is that the IM-mechanism measures the *novelty change* of the internal representation of the pattern, rather than its level: this can happen if the pattern is experienced multiple times and the source-component improves its representation each time. Typical functions played by the target-component, possibly coincident with the source-component, involve storing/re-coding of novel items, directing attention to novel items, driving their physical exploration, or supporting goal formation.

PB-IMs are based on a *predictor* source-component that predicts patterns on the basis of other patterns. In particular, the predictor receives as input a pattern, and possibly the agent's action, and on this basis predicts a target pattern in a future time. The "future time" involves a time range where the target item should happen, but predictions can also be "in space", as in this example: "given that I see a tree, I predict to see an apple if I look up 15 cm". The IM-mechanism of PB-IMs performs a measurement of the source-component (predictor) knowledge based on a two-step process: first the predictor predicts the target pattern on the basis of an input pattern, and possibly of the agent's action, and then the mechanism compares the prediction with the target pattern to compute the size of their mismatch - the *prediction error*. Another possibility is that the measure involves the *prediction error improvement* (change), rather than the *prediction error* (level), based on monitoring how the prediction error evolves in time. Typical functions played by the target-component, possibly coincident with the source-component, involve improving predictions, directing attention to unpredicted items, driving their physical exploration, and forming goals.

**Table 2**. The three classes of (e)IMs. Partially based on Barto et al. (2013).

|  | **Novelty-based IMs** | **Prediction-based IMs** | **Competence-based IMs** |
|---|---|---|---|
| **Source-component** | Memory component (pattern magazine) | Predictor (forward model) | Skill (inverse model) |
| **Function of the source-component** | Pattern storing and re-coding | Prediction of patterns based on other patterns | Action selection |
| **Type of knowledge-measured by the IM-mechanism** | How well-represented is the item in memory, or how much did its representation improve? How many times has the item been observed? | What is the prediction error, or the prediction error change? | How efficient/effective is the skill to accomplish the task/goal? |
| **Processes involved in the measurement of the IM-mechanism** | One process: memory check | Two processes: (a) prediction; (b) comparison of prediction with data | Multiple processes: iterated perception-action performance, check of the success |
| **Typical functions of the target-component** | - Store/re-code new items<br>- Direct attention<br>- Drive physical exploration<br>- Support goal formation | - Improve predictions<br>- Drive physical exploration<br>- Direct attention<br>- Support goal formation | - Speed up the learning of multiple skills |

CB-IMs assume the existence of tasks/goals, and are based on a *skill* source-component that can accomplish the tasks/goals (e.g., within a given period of time, the *trial*). The skill is a close- or open-loop controller (e.g., a dynamic movement primitive, a policy, or an option) potentially able to solve the task/achieve the goal. The IM-mechanism of CB-IMs performs a measurement of the knowledge of the source-component that involves a multi-step process: (a) the skill acts to accomplish the task/goal, possibly based on multiple sensorimotor steps; (b) its *competence level* is measured, e.g. in terms of the amount of reward collected during the trial, or in terms of goal achievement, or in terms of distance between the achieved state and the goal, etc. Another possibility is that the IM-mechanism actually measure the *competence improvement*, rather than the competence level, based on the monitoring of the performance at multiple times. CB-IMs are particularly important in cases where *multiple skills* for accomplishing different tasks/goals have to be learned. In this respect, typical functions played by the sklls (target-component), usually coincident with the source-component, is the selection of the skill/goal to learn to speed up the training of all skills/goals.

Note that the definition of CB-IMs assumes the existence of tasks/goals. This is a critical aspect of CB-IMs because open-ended learning agents should be able to autonomously generate or discover such tasks/goals as these are a major means to learn skills in an incremental fashion (Mirolli and Baldassarre 2013). Many oIMs considered in the following sections can be used to support such self-generation/discovery of tasks/goals.

# Extrinsic and intrinsic motivations: behaviour and brain

This section gives some examples of how IMs, in particular eIMs, are used in brain and behaviour of organisms. Before doing this, however, it gives an example on the brain and behaviour regulates EMs in relation to food ingestion as this allows a comparison of the mode of functioning of EMs vs. IMs. This example in particular shows how EMs "measure material resources" while eIMs "measure knowledge". The brain mechanisms supporting EMs are by far evolutionarily more ancient, more complex, and more studied than those supporting eIMs. For this reason, IM mechanisms often "evolutionary hijacked" existing brain structures emerged to support EM mechanisms.

## Extrinsic motivations

Animal brain has a complex information exchange not only with muscles and external senses of body (in particular hearing and vision) but also with the internal visceral body (Parisi 2004). For example, this is particularly apparent for food that is the main source of energy and body components and hence is paramount for survival and reproduction. *Hypothalamus* is a major hub of body homeostatic regulations and plays an important role in food intake regulation (Elmquist et al. 2005; Mirolli et al. 2010). Visceral body informs the hypothalamus on energy levels via a neural pathway involving the *nucleus of the solitary tract* and the *parabrachial nucleus*. Hypothalamus itself contains specialized glucose-sensitive neurons, is sensitive to the *leptine hormone* which grows when excessive adipose tissue in body, and is sensitive to gut hormones. Based on this information, the arcuate nucleus, and other nuclei of the lateral hypothalamus, directly control body metabolism (e.g., the conversion from sugar to fat and vice versa) by regulating the autonomic and endocrine systems, and also by controlling innate behaviours (e.g., feeding). Hypothalamus also activates, both directly and indirectly via the *peduncolopontine nucleus*, the *ventral tegmental area* producing *dopamine* (Geisler et al., 2007). Dopamine (Shultz 2002), especially when phasic (i.e. represented by a short intense activation of dopamine neurons) is one of the most important neuromodulators involved causing learning processes in *basal ganglia* and motor areas of the *frontal cortex,* the brain system underlying trial-and-error learning processes corresponding to RL.

There is a wide literature on the role of phasic dopamine for the learning of behaviour (for a review from a computational perspective, see Sutton and Barto 2018), fueled by one of the most fertile collaborations between neuroscience and and modelling of brain and behaviour in particular based on RL models. The key ideas of this link are captured by the *actor-critic RL model* (Barto 1995). This model is based on two components. The first component is the *actor* that learns to produce stimulus-response associations (*policy*). The second component is the *critic* that learns to predict future rewards and on this basis produces a step-by-step reward prediction error. This error, called *temporal difference error* (*TD-error*), indicates how much the actor performed better or worse than average in the last visited state. It is proposed that the actor and critic processes might actually be implemented in the brain, in particular the actor might correspond to a part of the basal ganglia called *striosomes*, the critic might correspond to another part of the basal-ganglia called *matriosomes*, and the TD-error might correspond to phasic dopamine signals (Houk and Barto 1995).

Aside the sensors detecting "material resources" *inside* the body, organisms often possess other sensors located *on the surface* of the body to detect the availability of resources in the reachable surrounding space. Smell and taste are examples of these sensors. These sensors are activated not only just before food is ingested, but also in anticipation of such an event, for example by the odours released by food in the air (reptiles "taste" odours in the air through typical tongue rhythmic movements). Also these sensors send information to brain centres involved in producing primary learning signals (Kandel et al. 2000). Based on this information, the brain also learns to detect the presence of resources in the environment through other distal sensors, for example hearing and vision. These processes, producing secondary learning signals, rely on the associative processes of amygdala representing an important brain hub for EMs (Mirolli et al. 2010).

## Intrinsic motivations

This section considers some biological systems that can be considered specific instances of the three classes of IM systems presented in the previous section. It also briefly mentions some models proposed to study such systems.

An important example of NB-IMs pivots on the *hippocampus*, a phylogenetically ancient cortex that plays a crucial role in the acquisition and consolidation of memories thanks to its widespread connections to most neocortex. The hippocampus is part of a brain system (Lisman and Grace 2005) that nicely reflects the organisation of such type of IMs. The hippocampus is highly activated by the perception of unfamiliar items, in particular novel objects or novel spatial arrangements of familiar objects (Kumaran, 2007). When it detects novel stimuli/configurations, the hippocampus activates the dopaminergic neurons of the *ventral tegmental area* that send dopamine to the hippocampus itself, and this favours the formation of memories of the new items. In this case, the source-component is the hippocampus, storing memories of precepts. The IM-mechanism is internal to the hippocampus, proposed to highly respond to novel stimuli as it compares stimuli with existing memories ("comparator hypothesis"), and on this basis to trigger dopamine production. The target-component is the hippocampus itself where the formation of the memory of new stimuli is facilitated by dopamine. A second example of NB-IM is related to another neuromodulator, *noradrenaline*, produced by the *locus coeruleus* in the brainstem. In rats, locus coeruleus activates in correspondence to novel stimuli and objects encountered during free exploration and then habituates after few experiences (Sara 1994), thus possibly influencing learning processes in cortical areas targeted by noradrenaline. The following section will discuss various machine learning models pivoting on mechanisms for novelty detection.

The most theoretically articulated and empirically supported theory on PB-IM is related to *superior colliculus* and phasic dopamine (Redgrave and Gurney, 2006). The theory proposes that *unexpected events*, e.g. changes of the environment caused by actions, activate the superior colliculus, an area of the midbrain that receives input from the retina and plays a key role in saccade generation. The superior colliculus causes phasic dopamine bursts of the *substantia nigra compacta* and the produced dopamine reaches basal ganglia few milliseconds after the stimulus. Here the dopamine signal causes the association between the sensations preceding the action and the performed action that possibly caused the stimulus. This process allows the organism to refine the action causing the effect. With repeated experiences, the superior colliculus response is progressively inhibited. After being formed, the skill can be recalled for execution if its outcome becomes desirable (e.g., on the basis of extrinsic motivations). Here the source-component is the neural system formed by the superior colliculus bursting to changes, and the syb-system that allows the anticipation and progressive inhibition of the superior colliculus, which has not yet been clearly identified (a hypothesis involves the basal ganglia themselves); the IM-mechanism is the stimulus prediction error causing the dopamine burst; and the target-component are the basal ganglia learning the skill.

Marraffa et al. (2012) proposed a model that captures some key elements of the latter system proposing a camera robot that learns to look the stimuli on a screen as in a target empirical experiment with babies (Want et al. 2012). The model self-generates a reward signal, similar to what would be done by the superior colliculus, when a new image suddenly appears at the centre of the screen as an effect of the fact that the system looks a certain target stimulus. This allows the model to learn the skill to visually navigate the screen to reliably cause the appearance of the image. Based on these processes, the model succeed to reproduce some of the effects found in the target experiment, for example the fact that after some time the children start to produce anticipatory saccades to the location where the image will appear.

There is not a complete example for CB-IM, but empirical evidence suggests some elements that might help to build such an example. The previous section has reviewed the brain machinery pivoting on basal ganglia and the dopamine system, studied with actor-critic models, that allows the acquisition of skills based on extrinsic rewards. If one assumes that the brain is able to autonomously produce *intrinsic rewards* from *self-generated tasks and goals* (discussed below), and to associate to their accomplishment an *intrinsic reward*, the rest of the machinery needed to implement a whole CB-IM system might be formed by the brain elements implementing the actor-critic functions. Indeed, as we have seen, the TD-error of the critic measures the fact the actor selected an action "better/worse than usual", so it can be seen as the IM-mechanism measuring the competence improvement of the actor (the source-component) used to train the actor itself to acquire the skill (target-component). It has been shown that indeed some brain areas of prefrontal cortex respond to the successful achievement of subgoals in extrinsic tasks and generate the production of a dopaminergic reward signal (Ribas-Fernades et al. 2011). So it might be hypothesised that also the accomplishment of intrinsic tasks/goals generates (intrinsic) rewards, and these might allow the actor-critic brain machinery to acquire the skills needed for such accomplishment.

The computational viability of such hypothesis has been shown with a model (Baldassarre et al., 2013) that reproduces the main structures of basal ganglia and cortex involved in the (self-)generation and encoding of goals, and the learning of the skills to accomplish them, driven by intrinsic rewards. The model allows an abstract agent to select actions to explore a "mechatronic board" having buttons that if touched produce different effects such as the opening of some boxes. Other models (Fiore et al. 2014; Mannella et al., 2016) show how IMs and EMs can largely rely on the same brain machinery. In particular, Fiore et al. (2014) shows how the same scenario of the mechatronic board, this time involving extrinsic rewards, is faced on the basis of the same brain skill learning machinery as in Baldassarre et al. (2013), but this time the primary and secondary rewards are managed by amygdala (Mannella et al. 2016; Mirolli et al. 2010).

# Cognitive robotics and machine learning models

This section considers the main functions that can be supported by IMs through the presentation of some computational models drawn from the robotics and machine learning literature. The section in particular focuses on how IMs serve the acquisition of the overall capacity of agents to interact in the world to modify it (Mirolli and Baldassarre 2013). This focus leads to consider in particular the relation between IMs and RL, the learning paradigm most closely related to the acquisition of the capacity to act in the world. Given this focus, the IM functions considered here are these (Figure 2): (a) the accomplishment of sparse extrinsic rewards; (b) the self-generation of goals; (c) the acquisition of skills, either as policies per se or as policies linked to goals. These functions are in particular accomplished through processes that strongly rely on IM mechanisms alongside other mechanisms; these latter processes are (a) exploration; (b) goal sampling, imagination, or "marking"; (c) and the autonomous selection of skills to learn. The model examples refer not only to robots but also to embodied agents acting in simulated scenarios as the proposed mechanisms might also be employed in robots.

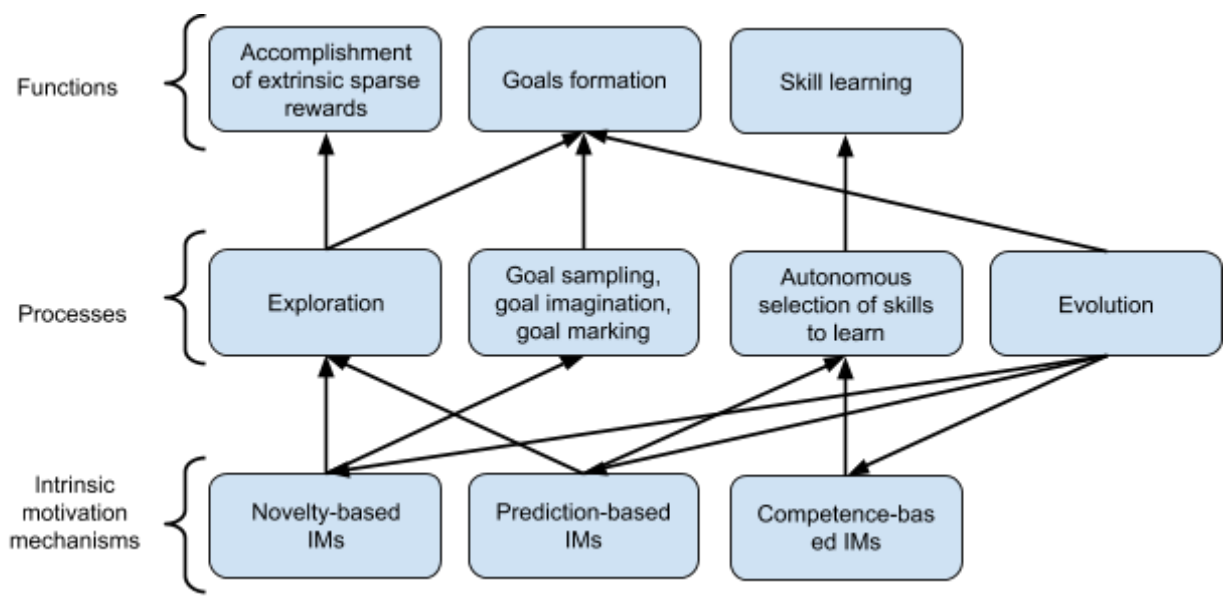

**Figure 2**: Some important functions that can be accomplished through some processes that strongly rely on IM mechanisms (with the exception of evolution).

The first main function of IMs is to support the solution of RL tasks involving *sparse* extrinsic rewards, i.e. rewards that are encountered rarely if the agent explores the environment randomly. Sparse rewards challenge learning agents as they can be experienced only after the performance of a long sequence of actions and so they furnish a very weak guidance for training. For example, imagine a camera-arm robot with no initial motor skills getting rewarded only for succeeding to grasp and lift an object with random movements: in this case, it is almost impossible that a random exploration leads to get the reward and support learning. IMs can be very useful to solve tasks involving sparse rewards as they can enhance the *exploration* of the environment through which the agent searches the reward. Standard exploration methods, such as the *ε-greedy* exploration (the agent selects a random action with a probability *ε,* and the best policy otherwise) and the *Boltzmann-distribution* exploration (the possible actions are selected on the basis of a *soft-max* function of their expected reward returns), are not adequate to face sparse-reward tasks. Various approaches have hence been proposed to produce a more effective exploration of the environment. A popular approach to foster exploration is based on NB-IMs. The idea is that the agent is attracted by states that it visited a few times and tends to move away from familiar states. An extra-reward (*novelty bonus*) could be given to the agent for making novel states attractive (Brafman and Tennenholtz 2002; Kakade and Dayan 2002). A nice property of novelty bonuses, and in general of IMs used to foster the pursuit of extrinsic rewards, is that since IMs have a transient nature they tend to not affect the final policy acquired to maximise the final extrinsic reward.

A relevant class of methods using novelty to foster exploration in the search of extrinsic rewards are based on state novelty measured as number of times that a state is encountered (Bellemare et al. 2016). In particular, these methods use density models to compute a pseudo-count of the times in which states are visited, based on the generalisation of the counts for similar states. The method was successfully applied to control agents able to solve the Atari game Montezuma Revenge involving a highly sparse reward. Another example is presented in Burda et al. (2018), using a different mechanism to compute the novelty of states. Here a random network is used to re-code the state observations (images), and a second "copy" network is trained with supervised learning to mimic such network (same input; desired output as the random network). The idea here is that when states become more familiar the error of the copy network decreases. The model is used to improve performance with Atari games, especially those involving sparse rewards.

Exploration to pursue extrinsic goals could be also pursued through PB-IMs. PB-IMs can rely on the *prediction error* (Schmidhuber 1991), or the *prediction error improvement* (Schmidhuber 1991a), of a predictor network (i.e. a world model predicting the next state on the basis of the current state and possibly the planned action). The prediction error has the disadvantage, if used by a IM-mechanism, that it does not fade away in stochastic words. This problem is solved by the prediction error improvement, although at the cost of having a noisy and slow-adjusting signal. In the initial models using this strategy, the predictor was used both as the source-component and as the target-component, meaning that the function of the used IM was to train the predictor itself. The same IM mechanism can however be also used to foster exploration to accomplish extrinsic tasks involving sparse rewards. An example of model doing this is presented in Pathack et al. (2017). Here a forward model is used to produce a prediction error used as intrinsic reward to train a RL agent to solve video games such as Mario Bross involving sparse extrinsic rewards. Interestingly, here the forward model does not take as input the observations of states but rather their encoding obtained by training an inverse model that gets as input a pair of succeeding states and learns to predict as output the action causing the transition between the pair. The encoding of the states is given by the intermediate layers of the network implementing the inverse model trained in a supervised fashion. The attempt of the inverse model to predict the action leads it to form encodings of states that have features that only covariate with the features of the world states that are affected by the agent's action, rather than with irrelevant state features not caused by the agent's action.

A very interesting function for which IMs can be used is related to the acquisition of multiple sensorimotor skills that might be later used to accomplish other intrinsic tasks, or extrinsic tasks, in particular within a hierarchical RL framework where behaviour is chunked into *options* (Sutton et al. 1999). Here we consider the goal-based version of options where each option involves (Barto et al. 2004, Singh et al. 2005): (a) a termination condition associated with the accomplishment of a *goal*; (b) an action policy indicating the primitive actions to select in correspondence to different states of the world; (c) (possibly) an initiation set encompassing the states from which if executed the policy is able to accomplish the goal. A goal is a representation of a set of world states that, if re-activated internally, drives the agent to act in the world so that the world assumes one of those states. There are various types of goals, for example goals as states of the world, goals as trajectories of states, avoidance goals, maintenance goals, etc. (Merrick et al. 2016). Here the focus is only on goals as states as this is the most important type of goals, but many considerations presented here could be extended to other types of goals. Goals can have different levels of abstraction and can involve own body (Mannella et al. 2018, Hoffmann et al. 2010) or the external environment (e.g., Santucci et al. 2016), or their relation (Kulkarni et al. 2016), possibly also involving social aspects (Acevedo-Valle et al. 2018).

There are multiple functions, supported by IMs, that are important for learning repertoires of multiple skills. These functions are all related to the fact that, during intrinsic autonomous learning, open-ended learning agents should learn as many skills as possible, and as well as possible, so that they can later best solve extrinsic tasks/goals. Here four of those functions are considered: (a) the autonomous generation of goals; (b) the coverage of the widest possible part of the goal space (*goal exploration*); (c) the generation of the reward for learning the policy of the single option; (d) the support of the progressive learning of skills, from easy to difficult, so to speed-up their acquisition.

The function of goal formation is related to the fact that during the intrinsic phase open-ended learning agents are not given any task to solve and so they should autonomously self-generate tasks/goals guiding the acquisition of the related skills. Note that although goal-formation is extremely important for open-ended learning, and various methods supporting it involve eIMs (Mirolli and Baldassarre 2013), it often also involves other mechanisms different from the elements of eIMsl (source-component, IM-mechanism, and target-component). These additional mechanisms are here considered to belong to oIMs: further investigations are needed to understand if these have common underlying factors and if and how they are related to eIMs. Various methods have been proposed to support the autonomous generation of goals and some relevant ones are now considered.

*Goal formation by sampling.* When the goal-space is given, e.g. it is formed by the posture angles of a robot or the x-y positions of an object on a table, goals can be sampled on the basis of their skill-learnability. For example, *goal babbling* (Rolf et al. 2005) allows a robot to self-generate posture goals that facilitate the learning of a coherent inverse model by maximising the end-effector displacement, which favours the exploration of novel goals, while minimising the posture change, which favours the learning of regular vs. awkward postures among the many possible (redundant) postures. The approach has been later extended in various directions, for example to learn multiple models in parallel (from end-effector position space to joint space, and from the joint space to the motor space) on the basis of an associative radial basis function network incrementally expanded on the basis of novel experiences (Rayyes and Steil 2019).

The goal-space might not be given to the agent, but form a subspace of the state- or observation-space to be actively searched. In this case goal sampling is not possible, especially if the subspace is small with respect to the whole space, and so the agent has to actively discovered by the agent. Consider for example an observation space formed by images: in this case, the agent has to actively discover the image goals that it might actually achieve with its actions within the whole huge space formed by all possible images corresponding to all combinations of the pixel values. Now some approaches usable for that purpose are considered.

*Mechanisms for goal marking.* A number of models have proposed specific mechanisms to "mark", i.e. recognise as goals, experienced states or observations. These models do not have the features of eIMs but can support open-ended learning via the formation of goals and the learning of the related skills, so they can be considered as oIMs. A classic approach is the one for marking as goals the experienced states of the world that represent *bottlenecks* (McGovern and Barto 2001), i.e. nodal conditions that are often traversed when solving multiple extrinsic tasks (e.g., doorways when navigating a house). Another model proposed to form goals corresponding to *salient events*, for example a change of light or sound (Barto et al. 2004; Singh et al. 2005). Another set of models used a strategy to mark as goals the observations that follow the production of *novel changes* caused by the agent's actions in the environment (Santucci et al. 2016; Mannella et al. 2018). The idea behind this approach is that what robots (and organisms) ultimately should do during intrinsic learning is to become able to change the world at will, so the observations that follow a change caused by their actions indicates a potential for doing this. The novelty of the changes guarantees that the goal has not been already formed. If changes in the world can happen also independently of the agent's action, additional mechanisms are needed to allow the agent to identify those changes that depend on its action (Sperati and Baldassarre 2018). Another approach forms goals when a particular relation between couples of elements takes place, e.g. the agent picks up a key in an Atari game (Kulkarni et al. 2016). A different approach (Zhao et al. 2012) proposes a motorised looking camera to use RL to acquire various behaviours in the field of active vision (Ballard 1991; Ognibene and Baldassarre 2015), such as vergence control, by using as reward the *accuracy of the reconstruction* of images of a sparse-coding component (Olshausen and Field 1996). Here the model seems to have one source-component (the sparse-coding component) but a double function (two target-components), one directed to acquire useful behaviours (RL component), as vergence control, and another directed to favour the acquisition of good internal representations of percepts (two target-components). Another interesting approach is *empowerment* (Klyubin et al. 2005), based on information theory, that can be used to assign to each given world state a value that represents the variety of different outcome states that the agent can achieve with its actions from the given state. States with high empowerment can be used as target states, in particular their empowerment value can be directly used as reward to drive the learning of skills in the absence of tasks and extrinsic rewards (Jung et al. 2011). The overall concept of empowerment has a scope and relevance for open-ended learning that is broader than the one discussed here, but discussing this would go beyond the scope of this contribution. Der and Martius (2015) propose another approach that uses emergent properties of the environment-body-controller dynamics to acquire interesting motor skills without the use of extrinsic rewards. The skills are in particular acquired on the basis of a simple two-layer neural-network sensorimotor controller whose connection weights are trained through a *Differential Extrinsic Plasticity* (DEP) rule that is derived from *Differential Hebbian Learning* (Zappacosta et al. 2018), which updates connection weights if the pre-synaptic neuron changes correlate with the post-synaptic neuron changes, but is also sensitive to physical changes affecting the neurons and caused by the environment.

*Goal discovery by exploration*. Another strategy searches goals based on the idea that similar goals involve similar skills, and so the performance of noisy variants of the already discovered skills might possibly lead to new viable goals. This strategy was first used in a model (*skill babbling*; Reinhart 2017) to control an arm robot learning to displace an object in the 3D space. The model forms clusters of similar goals and discovers new goals by performing noisy versions of the skills corresponding to the centroid goals of the clusters, in particular focussing on the most recent clusters. The *active goal manifold exploration* model (AGME; Cartoni and Baldassarre, 2018) goes beyond the discretisation of the goal-space in clusters by actively constructing a homogeneous representation of the goal manifold hidden in the observation space (from posture angles to image pixels). To this purpose, the model builds a distance-based graph of the discovered goals, selects goals that have a higher average distance from other discovered goals, generates perturbed versions of the policies associated with such goals, and performs them so possibly discovering new goals. Another model (the *quality diversity algorithm*, Kim et al. 2019) learns a repertoire of behaviours and goals by searching behaviours that are different (novel) with respect to the already learned behaviours. The algorithm is for example used to allow a humanoid robot to acquire the skills to throw a ball into a basket located in many possible different positions (goals) on the floor. Another approach, called Hindsight Experience Replay (*HER*; Adrychowicz et al. 2017), exploits the unexpected outcome of policies to discover new goals ("unexpected" as different from the pursued goal). The approach is shown to be very effective to incrementally discover new goals, e.g. for the manipulation of objects in a simulated camera-arm-gripper robot.

*Goal formation by imagination*. Another related strategy tries to discover goals by first imagining them and then by pursuing them in the environment to check for their existence (or to discover similar goals). For example, the *reinforcement learning with imagined goals* model (RIG; Nair et al, 2018), tested with a robot arm moving objects on a table, uses a generative model (a variational autoencoder; Kingma and Welling 2013) to first learn an internal compact representation of goals by randomly exploring the environment, and then to "imagine" other possible goals whose skills are learned by RL. A later version of the model generates goals that have a high probability of being novel with respect to already learned goals by sampling them on the fringe of the distribution of the internal representation of the discovered goals (Pong et al. 2019). Incidentally, imagination is a relevant means not only to generate goals but also to formulate plans to achieve those goals by assembling other goals/skills (Seepanomwan et al. 2015, Hung et al. 2018, Jung et al. 2019, Tanneberg et al. 2019) possibly acquired with IMs: this is an interesting recent trend that aims to re-formulate some of the concepts relevant for higher level cognition elaborated by the classic symbolic planning literature (Russell and Norvig 2016), such as goals and planning, through neural-network representations (Baldassarre et al. 2001; Wayne et al. 2018).

*Selection of the skills/goals to train*. The literature on animal learning (Skinner 1953) and on staged children development (Piaget 1953) shows that learning progress is faster if it proceeds from easy to difficult tasks. This strategy can also be used with artificial systems by training them with a *curriculum* involving increasingly difficult tasks (Asada et al. 1996; Bengio et al. 2009). One of the most interesting uses of IMs allows open-ended learning agents to *autonomously* select increasingly difficult tasks to learn, in particular tasks posing the challenges suitable for the current learner's knowledge. These processes can be used for learning sets of given tasks or tasks autonomously generated through the algorithms seen above. Initially, PB-IMs were used to support the autonomous selection of tasks to learn (e.g. Singh et al. 2004; Oudeyer et al., 2007). Here the source-component was a predictor while the target-component was the skill to be trained. The agent in particular focussed its learning on skills causing the highest predictor error, or prediction error improvement, related to the prediction of the skill outcome. Successively, CB-IMs were shown to be more appropriate than PB-IMs for selecting the skills to train because the predictor of the PB-IMs might learn to predict the skill outcome too early or too late with respect to when the controller finishes to learn the skill; instead, CB-IMs directly measure the competence acquired by the different skills and so it returns a better information to select them (Santucci et al. 2013 compared these different IM mechanisms for task selection). CB-IMs can also be used when the conditions from which to accomplish the tasks are variable, a situation implying the additional challenge that competence differs for the different conditions; or when the agent might decide to select a skill, notwithstanding it is already well trained, to accomplish its goal when this is also a precondition for learning other skills (Santucci et al. 2019). IMs can also guide the progressive learning of increasingly difficult tasks represented at multiple levels of abstraction, e.g. in robots learning to interact with different objects (Ugur and Piater 2016). In all these models, and contrary to what was done in early models using PB-IMs, the selected skills are trained with RL through a *pseudo-reward* equal to one when the goal is accomplished and to zero otherwise: this is a clearer signal to use for this purpose as IM signals fade away with learning.

*Evolution*. Tasks/goals could be also generated autonomously through evolutionary algorithms that search for them on the basis of an objective function (fitness) based on the success in solving extrinsic tasks through the skills learned on the basis of the self-generated tasks/goals. Schembri et al. (2007) proposed a first model to do so in a population of RL simulated robots that in a first "childhood phase" used evolved intrinsic reward functions to learn skills. In a later "adulthood phase" the robots learned to compose the acquired skills to accomplish extrinsic tasks, requiring to follow coloured paths on the floor: these extrinsic tasks produced the fitness for the genetic algorithm. Singh et al. (2010) used an algorithm equivalent to evolution to search reward functions of RL agents engaged in searching for food in a grid word. The found reward functions with the highest score led the agents to be rewarded not only for searching for food but also for "opening boxes" where food was hidden. The model was used to argue the existence of a continuum between EMs and IMs, rather than a distinction between them, as the two differ only for their distance from the final extrinsic reward. The view proposed here on eIMs should clearly distinguish them from EMs as eIMs are based on an IM-mechanism that measures the *knowledge* of a component of the controller whereas EMs are based on the measure of *material resources* in the body or in the environment. It is however true that in the case of evolved oIMs that support the formation of goals and skills, as in the models above, a continuum with EMs can be seen since the criterion of the knowledge-measurement mechanism of eIMs is missing. An additional crucial problem for open-ended learning that can be tackled with evolutionary approaches is the following one: which goals/skills should be acquired, among the possible ones, to later best solve several different extrinsic tasks in a given domain? This problem was faced in Del Verme et al. (2019) that used a genetic algorithm to search goals/skills that were optimal for the solution of tasks drawn from a certain distribution of possible tasks in a given domain. The work showed how the optimal goals and skills depended on the time budget that the agent had in order to solve the extrinsic tasks and on the physical regularities of the environment, thus showing that "fixed" mechanisms for goal generation as those seen above might lead to suboptimal solutions. Importantly, evolutionary approaches might also be used to evolve the IM-mechanisms altogether, as hinted by the arrows in Figure 2 departing from the "evolution box" (Salgado et al. 2016). A problem with this objective is that it is computationally very expensive.

# Conclusion and open challenges

This section closes the paper by illustrating some of the open issues of open-ended learning and intrinsic motivations. Some of the open issues involve neuroscience and psychology. A first important objective, shared with other topics of investigation, would be to more often formalise theories on brain and behaviour as computational models rather than only as verbal models (Naselaris et al. 2018). This would have the advantages of specifying theories, of producing quantitative predictions testable in empirical experiments, and of favouring the progressive integration of psychological/neuroscientific knowledge now scattered in multiple parts related to different experimental paradigms and topics. For example, one might investigate the biological and behavioural basis of the distinction between novelty and surprise intrinsic motivations, which have been distinguished within the computational literature but are still confused within the empirical research (Barto et al. 2013). An important overall challenge related to the study of intrinsic motivations is that these are elicited only when the experimental conditions pose challenges that are not too easy nor too difficult for the participant, and this strongly depends on the specific experiences of each different participant. One way to face this problem is to create unusual experimental conditions that one is sure are novel for all participants and so will be engaging for all of them (Taffoni et al. 2014), but the challenge of tuning the right level of challenge for all participants still remains. Another challenge involves autonomous goal generation processes. We have seen that these processes are pivotal for open-ended learning. This topic, with few exceptions (e.g. Ribas-Fernandes et al. 2011), is now largely neglected by neuroscience and psychology notwithstanding its dramatic importance. A reason is probably the difficulty of its study as it does not allow the exploitation of the usual empirical paradigms based on extrinsic tasks that facilitate having tight controlled empirical conditions. Another related hard obstacle is the difficulty of empirically measuring and manipulating the goals of participants since goals are a very complex construct ("internal representations that, if activated internally, drive behaviour and learning"). One way to manipulate goals is for example the *devaluation paradigm* (Kenward 2009; Mannella et al. 2016): this gives an idea of the level of complexity of the experimental paradigms needed to manipulate goals. A construct related to goals that is easier to study than them is represented by *sensorimotor contingencies* (Jacquey et al. 2019; Polizzi di Sorrentino et al. 2014), but these do not fully capture the richness and cognitive importance of goals. Notwithstanding these and other difficulties the empirical study of intrinsic motivations and of the autonomous formation of goals is strongly needed as these processes are at the basis of humans' curiosity, creativity, science, and art, and also crucial for human well-being (Ryan and Deci, 2000).

On the computational side, the problem of open-ended learning is still unsolved as shown by the fact that we still do not have robots able to undergo a truly open-ended learning experience leading to an unbounded accumulation of knowledge and skills. This might depend on multiple factors. On the side of goal formation, we have various mechanisms for the autonomous generation of goals, but none of them has so far revealed as resolutive. Future work should hence aim to further develop the different approaches, and also to evaluate which ones can scale up to support the acquisition of increasingly complex skills. Indeed, all known approaches have limitations: goal sampling (e.g. Rolf et al. 2005) can only be applied when the goal space is known; goal formation based on mechanisms such as bottlenecks (e.g. McGovern and Barto 2001), novel environment changes caused by actions (e.g. Santucci et al. 2016), empowerment (Klyubin et al. 2005), and Differential Extrinsic Plasticity (Der and Martius 2015), have to be tested for their applicability to a wider range of domains; goal exploration (e.g. Reinhart 2018; Cartoni and Baldassarre, 2018), and goal discovery by imagination (e.g. Pong et al. 2019) have still to show that they can scale up larger sets goals. Alternatively, new approaches might be proposed, also possibly based on the integration of previous approaches. The autonomous selection of the skills to train, based on competence-based intrinsic motivations (e.g. Santucci et al. 2016) is reliable but assumes discrete goals which require the development of suitable solutions to be applied to continuous goal spaces. Moreover, systems working with discrete goals solve extrinsic problems based on planning/search methods (e.g. Baldassarre et al. 2018) which require that the number of learned goal/skills is limited to be efficient. This problem might be faced with evolutionary methods that search few efficient skills that lead to the optimal solution of multiple tasks drawn from the possible ones in a certain domain (e.g. Schembri et al. 2007; Singh et al. 2010; Del Verme et al. 2019). However, the computational burden of these approaches is so high that they are still only applicable to simple domains given the computational power of current computers.

# Acknowledgment

This research received funding from the European Union under: the 7th Research Framework Program, Grant Agreement no. 231722, Project "IM-CLeVeR - Intrinsically Motivated Cumulative Learning Versatile Robots"; and Horizon 2020 Research and Innovation Program, Grant Agreement no. 713010, Project "GOAL-Robots - Goal-based Open-ended Autonomous Learning Robots".

# References

Acevedo-Valle, J. M., Hafner, V. V. & Angulo, C. (2018). Social Reinforcement in Artificial Prelinguistic Development: A Study Using Intrinsically Motivated Exploration Architectures. IEEE Transactions on Cognitive and Developmental Systems. DOI: 10.1109/TCDS.2018.2883249.

Andrychowicz, M., Wolski, F., Ray, A., Schneider, J., Fong, R., Welinder, P., McGrew, B., Tobin, J., Abbeel, P. & Zaremba, W., (2017). Hindsight experience replay. arXiv: 1707.01495v3.

Asada, M., Noda, S., Tawaratsumida, S. & Hosoda, K. (1996). Purposive behavior acquisition for a real robot by vision-based reinforcement learning. Machine learning, 23, 279-303.

Baldassarre, G., Lord, W., Granato, G. & Santucci, V. G. (2019). An embodied agent learning affordances with intrinsic motivations and solving extrinsic tasks with attention and one-step planning. Frontiers in Neurorobotics, 13, 45.

Baldassarre, G., Mannella, F., Fiore, V. G., Redgrave, P., Gurney, K. & Mirolli, M. (2013). Intrinsically motivated action-outcome learning and goal-based action recall: A system-level bio-constrained computational model. Neural Networks, 41, 168-187.

Baldassarre, G. (2001). Planning with neural networks and reinforcement learning. e1-191. PhD thesis. Department of Computer Science, University of Essex.

Baldassarre, G. (2011). What are intrinsic motivations? A biological perspective. In Cangelosi, A., Triesch, J., Fasel, I., Rohlfing, K., Nori, F., Oudeyer, P.-Y., Schlesinger, M. & Nagai, Y. (eds.), Proceedings of the International Conference on Development and Learning and Epigenetic Robotics (ICDL-EpiRob-2011), e1-8.

Baldassarre, G. & Mirolli, M. (2013) (eds.). Intrinsically motivated learning in natural and artificial systems. Berlin: Springer-Verlag.

Ballard, D. H. (1991). Animate vision. Artificial intelligence, 48, 57-86.

Barto, A., Mirolli, M. & Baldassarre, G. (2013). Novelty or surprise? Frontiers in Psychology, 4, e1-15.

Barto, A. G., Singh, S. & Chentanez, N. (2004), Intrinsically Motivated Learning of Hierarchical Collections of Skills, in International Conference on Developmental Learning (ICDL-2004). 20-22 October 2004. La Jolla, California, USA.

Barto, A. G. (1995). Adaptive Critics and the Basal Ganglia. In Houk, J. C., Davids, J. L. & Beiser, D. G. (eds.), Models of Information Processing in the Basal Ganglia, 215. Boston, MA: The MIT Press.

Bellemare, M. G., Srinivasan, S., Ostrovski, G., Schaul, T., Saxton, D. & Munos, R. (2016). Unifying Count-Based Exploration and Intrinsic Motivation. arXiv: 1606.01868.

Bengio, Y., Louradour, J., Collobert, R. & Weston, J. (2009). Curriculum learning. Proceedings of the 26th annual international conference on machine learning, 41-48.

Berlyne, D. E. (1966), Curiosity and Exploration. Science, 143, 25-33.

Brafman, R. I. & Tennenholtz, M., (2002). R-max-a general polynomial time algorithm for near-optimal reinforcement learning. Journal of Machine Learning Research, 3, 213-231.

Burda, Y., Edwards, H., Storkey, A. & Klimov, O. (2018). Exploration by Random Network Distillation. arxiv: 1810.12894v1.

Cartoni, E. & Baldassarre, G. (2018). Autonomous discovery of the goal space to learn a parameterized skill. arXiv: 1805.07547v1.

Del Verme, M., Castro da Silva, B. & Baldassarre, G. (2019). Optimal Options for Multi-Task Reinforcement Learning Under Time Constraints. The International Conference on Reinforcement Learning and Decision Making (RLDM-2019), e1-5. 7-10 July 2019, Montréal, Canada.

Der, R. & Martius, G. (2015). Novel plasticity rule can explain the development of sensorimotor intelligence. Proceedings of the National Academy of Science, 112, e6224-6232.

Doncieux, S., Filliat, D., Díaz-Rodríguez, N., Hospedales, T., Duro, R., Coninx, A., Roijers, D. M., Girard, B., Perrin, N. & Sigaud, O. (2018). Open-ended learning: a conceptual framework based on representational redescription. Frontiers in Neurorobotics, 12, 59.

Doya, K. & Taniguchi, T. (2019). Toward evolutionary and developmental intelligence. Current Opinion in Behavioral Sciences, 29, 91-96.

Elmquist, J. K., Coppari, R., Balthasar, N., Ichinose, M. & Lowell, B. B. (2005), Identifying hypothalamic pathways controlling food intake, body weight, and glucose homeostasis. Journal of Comparative Neurology, 493(1), 63-71.

Fiore, V. G., Sperati, V., Mannella, F., Mirolli, M. Gurney, K., Firston, K., Dolan, R. J. & Baldassarre, G. (2014). Keep focussing: striatal dopamine multiple functions resolved in a single mechanism tested in a simulated humanoid robot. Frontiers in Psychology - Cognitive Science, 5, e1-17.

Geisler, S., Derst, C., Veh, R. W. & Zahm, D. S. (2007), Glutamatergic afferents of the ventral tegmental area in the rat. Journal of Neuroscience, 27(21), 5730-5743.

Goodfellow, I., Bengio, Y. & Courville, A. (2017). Deep Learning. Boston, MA: The MIT Press.

Harlow, H. F. (1950), Learning and satiation of response in intrinsically motivated complex puzzle performance by monkeys. Journal of Comparative and Physiological Psychology, 43, 289-294.

Hoffmann, M., Marques, H., Arieta, A., Sumioka, H., Lungarella, M. & Pfeifer, R. (2010). Body schema in robotics: a review. IEEE Transactions on Autonomous Mental Development, 2, 304-324.

Houk, J. C., Adams, J. L. & Barto, A. G. (1995). A model of how the basal ganglia generate and use neural signals that predict reinforcement. In Houk, J. C., Davids, J. L. & Beiser, D. G. (eds.), Models of Information Processing in the Basal Ganglia, 249-270. Boston, MA: The MIT Press.

Hull, C. L. (1943), Principles of Behavior, Appleton Century Crofts, New York, NY.

Hung, C.-C., Lillicrap, T., Abramson, J., Wu, Y., Mirza, M., Carnevale, F., Ahuja, A. & Wayne, G. (2018). Optimizing agent behavior over long time scales by transporting value. arXiv: 1810.06721v1.

Jacquey, L., Baldassarre, G., Santucci, V. G. & O'Regan, J. K. (2019). Sensorimotor contingencies as a key drive of development: from babies to robots. Frontiers in Neurorobotics, 13, 1-20.

Jung, M., Matsumoto, T. & Tani, J. (2019). Goal-directed behavior under variational predictive coding: dynamic organization of visual attention and working memory. arXiv: 1903.04932v1.

Jung, T., Polani, D. & Stone, P. (2011). Empowerment for continuous agent environment systems. Adaptive Behavior, 19, 16-39.

Kakade, S. & Dayan, P. (2002). Dopamine: generalization and bonuses. Neural Networks, 15, 549-559.

Kandel, E. R., Schwartz, J. H. & Jessel, T. M. (2000). Principles of Neural Science. McGraw-Hill, New York.

Kenward, B., Folke, S., Holmberg, J., Johansson, A. & Gredebäck, G. (2009). Goal directedness and decision making in infants. Developmental Psychology, 45, 809-819.

Kim, S., Coninx, A. & Doncieux, S. (2019). From exploration to control: learning object manipulation skills through novelty search and local adaptation. arXiv: 1901.00811v1.

Kingma, D. P. & Welling, M. (2013). Auto-Encoding Variational Bayes. Arxiv: 1312.6114v10.

Kish, G. B. (1955), Learning when the onset of illumination is used as the reinforcing stimulus. Journal of Comparative and Physiological Psychology, 48(4), 261-264.

Klyubin, A., Polani, D. & Nehaniv, C. (2005). All else being equal be empowered. European conference on Artificial Life (ECAL-2005), 744-753. 5-9 September 2005. Canterbury, UK,

Kulkarni, T. D., Narasimhan, K. R., Saeedi, A. & Tenenbaum, J. B. (2016). Hierarchical deep reinforcement learning: Integrating temporal abstraction and intrinsic motivation. arXiv: 1604.06057.

Kumaran, D. & Maguire, E. A. (2007), Which computational mechanisms operate in the hippocampus during novelty detection? Hippocampus, 17(9), 735-748.

Lieberman, D. A. (1993). Learning, Behaviour and Cognition. New York, Brooks/Cole.

Lisman, J. E. & Grace, A. A. (2005), The hippocampal-VTA loop: controlling the entry of information into long-term memory. Neuron, 46(5), 703-713.

Lungarella, M., Metta, G., Pfeifer, R. & Sandini, G., (2003). Developmental robotics: a survey. Connection Science, 15, 151-190.

Mannella, F., Mirolli, M. & Baldassarre, G. (2016). Goal-directed behavior and instrumental devaluation: A neural system-level computational model. Frontiers in Behavioral Neuroscience, 10, e1-27.

Mannella, F., Santucci, V. G., Eszter, S., Jacquey, L., O'Regan, K. J. & Baldassarre, G. (2018). Know your body through intrinsic goals. Frontiers in Neurorobotics, 12, e1-17.

Marr, D. & Poggio, T. (1976). From understanding computation to understanding neural circuitry. Technical Report AIM-357. Massachusetts Institute of Technology, Artificial Intelligence Laboratory.

Marraffa, R., Sperati, V., Caligiore, D., Triesch, J. & Baldassarre, G. (2012). A bio-inspired attention model of anticipation in gaze-contingency experiments with infants. In Movellan, J. & Schlesinger, M. (eds.), IEEE International Conference on Development and Learning-EpiRob 2012 (ICDL-EpiRob-2012), e1-8.

McGovern, A. & Barto, A. (2001). Automatic discovery of subgoals in reinforcement learning using diverse density. Technical report 8. Computer Science Department, University of Massachusetts.

Merrick, K., Siddique, N. & Rano, I. (2016). Experience-based generation of maintenance and achievement goals on a mobile robot. Journal of Behavioral Robotics, 7(1), 67-84.

Mirolli, M., Mannella, F. & Baldassarre, G. (2010), The roles of the amygdala in the affective regulation of body, brain and behaviour. Connection Science, 22(3), 215-245.

Mirolli, M. & Baldassarre, G. (2013). Functions and mechanisms of intrinsic motivations: The knowledge versus competence distinction. In Baldassarre, G. & Mirolli, M. (eds.), Intrinsically Motivated Learning in Natural and Artificial Systems, 49-72. Berlin: Springer-Verlag.

Nair, A., Pong, V., Dalal, M., Bahl, S., Lin, S. & Levine, S. (2018). Visual reinforcement learning with imagined goals. The Second Lifelong Learning: A Reinforcement Learning Approach Workshop (LLRLA-2018 at FAIM-2018). 14-15 July 2018. Stockholm, Sweden.

Naselaris, T., Bassett, D. S., Fletcher, A. K., Kording, K., Kriegeskorte, N., Nienborg, H., Poldrack, R. A., Shohamy, D. & Kay, K. (2018). Cognitive Computational Neuroscience: A new conference for an emerging discipline. Trends in Cognitive Sciences, 22, 365-367.

Ognibene, D. & Baldassarre, G. (2015). Ecological active vision: four bio-inspired principles to integrate bottom-up and adaptive top-down attention tested with a simple camera-arm robot. IEEE Transactions on Autonomous Mental Development, 7, 3-25.

Olshausen, B. A. & Field, D. J. (1996). Emergence of simple-cell receptive field properties by learning a sparse code for natural images. Nature, 381, 607-609.

Oudeyer, P. Y., Kaplan, F. & Hafner, V. V. (2007). Intrinsic motivation systems for autonomous mental development. IEEE Transactions on Evolutionary Computation, 11(2), 265-286.

Oudeyer, P.-Y. & Kaplan, F. (2007). What is intrinsic motivation? A typology of computational approaches. Frontiers in Neurorobotics, 1, 6.

Panksepp, J. (1998), Affective Neuroscience: The Foundations of Human and Animal Emotions. Oxford University Press, Oxford.

Parisi, D. (2004), Internal robotics. Connection Science, 16(4), 325-338.

Pathak, D., Agrawal, P., Efros, A. A. & Darrell, T. (2017). Curiosity-driven exploration by self-supervised prediction. arXiv: 1705.05363.

Piaget, J. (1953). The Origins of Intelligence in Children. London: Routledge and Kegan Paul.

Polizzi di Sorrentino, E., Sabbatini, G., Truppa, V., Bordonali, A., Taffoni, F., Formica, D., Baldassarre, G., Mirolli, M. & Guglielmelli Eugenio, V. E. (2014). Exploration and learning in capuchin monkeys (Sapajus spp.): the role of action-outcome contingencies. Animal Cognition, 17, 1081-1088.

Pong, V. H., Dalal, M., Lin, S., Nair, A., Bahl, S. & Levine, S. (2019). Skew-fit: State-covering self-supervised reinforcement learning. arXiv: 1903.03698v2.

Rayyes, R. & Steil, J. (2019). Online associative multi-stage goal babbling toward versatile learning of sensorimotor skills. The Joint IEEE 9th International Conference on Development and Learning and Epigenetic Robotics (ICDL-EpiRob2019), 327-334. 19-22 August 2019. Oslo, Norway.

Redgrave, P. & Gurney, K. (2006). The short-latency dopamine signal: a role in discovering novel actions? Nature Review Neuroscience, 7(12), 967-975.

Reinhart, R. F. (2017). Autonomous exploration of motor skills by skill babbling. Autonomous Robots, 41, 1521-1537.

Ribas-Fernandes, J. J. F., Solway, A., Diuk, C., McGuire, J. T., Barto, A. G., Niv, Y. & Botvinick, M. M. (2011). A neural signature of hierarchical reinforcement learning. Neuron, 71, 370-379.

Rolf, M., Steil, J. J. & Gienger, M. (2010). Goal babbling permits direct learning of inverse kinematics. IEEE Transactions on Autonomous Mental Development, 2, 216-229.

Russell, S. J. & Norvig, P. (2016). Artificial Intelligence: A Modern Approach. New York, Pearson Education.
Ryan, R. M. & Deci, E. L. (2000), Intrinsic and extrinsic motivations: Classic definitions and new directions. Contemporary Educational Psychology, 25, 54-67.
Ryan, R. M. & Deci, E. L. (2000). Self-determination theory and the facilitation of intrinsic motivation, social development, and well-being. American Psychologist, 55, 68-78.
Salgado, R., Prieto, A., Caamaño, P., Bellas, F. & Duro, R. J. (2016). MotivEn: Motivational engine with sub-goal identification for autonomous robots. IEEE Congress on Evolutionary Computation (CEC-2016), 4887-4894. 24-29 July 2016. Vancouver BC, Canada.
Santucci, V. G., Baldassarre, G. & Cartoni, E. (2019). Autonomous Reinforcement Learning of Multiple Interrelated Tasks. The 9th International Conference on Development and Learning and Epigenetic Robotics (ICDL-Epirob-2019). 19-22 August 2019. Oslo, Norway.
Santucci, V. G., Baldassarre, G. & Mirolli, M. (2013). Which is the best intrinsic motivation signal for learning multiple skills? Frontiers in Neurorobotics, 7, e1-14.
Santucci, V. G., Baldassarre, G. & Mirolli, M. (2016). GRAIL: A goal-discovering robotic architecture for intrinsically-motivated learning. IEEE Transactions on Cognitive and Developmental Systems, 8, 214-231.
Sara, S. J., Vankov, A. & Hervé, A. (1994), Locus coeruleus-evoked responses in behaving rats: a clue to the role of noradrenaline in memory. Brain Research Bulletin, 35(5-6), 457-465.
Schembri, M., Mirolli, M. & Baldassarre, G. (2007). Evolving childhood's length and learning parameters in an intrinsically motivated reinforcement learning robot. In Berthouze, L., Dhristiopher, G. P., Littman, M., Kozima, H. & Balkenius, C. (eds.). Proceedings of the Seventh International Conference on Epigenetic Robotics (EpiRob2007), 141-148. Lund University Cognitive Studies, 134.
Schmidhuber, J. (1991). A possibility of implementing curiosity and boredom in model-building neural controllers. In Meyer,J.-A. & Wilson, S. W. (eds.), Proceedings of the International Conference on Simulation of Adaptive Behavior: From Animals to Animats, 222-227. Boston, MA: The MIT Press.
Schmidhuber, J. (1991a), Curious model-building control systems. In Proceedings of the International Joint Conference on Neural Networks, pp. 1458-1463. 18-21 November 1991. Singapore.
Schmidhuber, J. (2010). Formal theory of creativity, fun, and intrinsic motivation (1990-2010). IEEE Transactions on Autonomous Mental Development, 2(3), 230-247.
Schultz, W. (2002). Getting formal with dopamine and reward. Neuron, 36(2), 241-263.
Seepanomwan, K., Caligiore, D., Cangelosi, A. & Baldassarre, G. (2015). Generalization, decision making, and embodiment effects in mental rotation: a neurorobotic architecture tested with a humanoid robot. Neural Networks, 72, 31-47.
Singh, S., Lewis, R. L., Barto, A. G. & Sorg, J. (2010). Intrinsically motivated reinforcement learning: An evolutionary perspective. IEEE Transactions on Autonomous Mental Development, 2(2), 70-82.
Singh S., Barto A. G. & Chentanez, N. (2005). Intrinsically motivated reinforcement learning. In L. K. Saul, Y. Weiss & L. Bottou (eds)., Advances in Neural Information Processing Systems 17: Proceedings of the 2004 Conference. Cambridge, MA: The MIT Press.
Skinner, B. F. (1938), The Behavior of Organisms. New York, NY: Appleton Century Crofts.
Skinner, B.F. (1953). Science and human behavior. Oxford, England: Macmillan.
Sperati, V. & Baldassarre, G. (2018). A bio-inspired model learning visual goals and attention skills through contingencies and intrinsic motivations. IEEE Transactions on Cognitive and Developmental Systems, 10, 326-344.

Sutton, R., Precup, D. & Singh, S. (1999). Between MDPs and semi-MDPs: A framework for temporal abstraction in reinforcement learning. Artificial Intelligence, 112, 181-211.

Sutton, R. S. & Barto, A. G. (2018). Reinforcement learning: an introduction. Boston, MA: The MIT Press.

Taffoni, F., Tamilia, E., Focaroli, V., Formica, D., Ricci, L., Di Pino, G., Baldassarre, G., Mirolli, M., Guglielmelli, E. & Keller, F. (2014). Development of goal-directed action selection guided by intrinsic motivations: an experiment with children. Experimental Brain Research, 232, 2167-2177.

Tanneberg, D., Peters, J. & Rueckert, E. (2019). Intrinsic motivation and mental replay enable efficient online adaptation in stochastic recurrent networks. Neural Networks, 109, 67-80.

Tinbergen, N. (1963), On aims and methods of ethology, Zeitschrift fur Tierpsychologie 20 (4), 410-433.

Ugur, E. & Piater, J. (2016). Emergent structuring of interdependent affordance learning tasks using intrinsic motivation and empirical feature selection. IEEE Transactions on Cognitive and Developmental Systems, 9, 328-340.

Vieira, N. H. & Nehmzow, U. (2007), Visual novelty detection with automatic scale selection, Robotics and Autonomous Systems 55, 693-701.

Wang, Q., Bolhuis, J., Rothkopf, C. A., Kolling, T., Knopf, M. & Triesch, J. (2012). Infants in Control: Rapid Anticipation of Action Outcomes in a Gaze-Contingent Paradigm. PLoS one, 7, e30884.

Wayne, G., Hung, C.-C., Amos, D., Mirza, M., Ahuja, A., Grabska-Barwinska, A., Rae, J., Mirowski, P., Leibo, J. Z., Santoro, A., Gemici, M., Reynolds, M., Harley, T., Abramson, J., Mohamed, S., Rezende, D., Saxton, D., Cain, A., Hillier, C., Silver, D., Kavukcuoglu, K., Botvinick, M., Hassabis, D. & Lillicrap, T. (2018). Unsupervised Predictive Memory in a Goal-Directed Agent. arXiv: 1803.10760v1.

Weng, J., McClelland, J., Pentland, A., Sporns, O., Stockman, I., Sur, M. & Thelen, E. (2001). Autonomous mental development by robots and animals. Science, 291, 599-600.

White, R. W. (1959), Motivation reconsidered: The concept of competence. Psychological Review, 66, 297-333.

Zappacosta, S., Mannella, F., Mirolli, M. & Baldassarre, G. (2018). General differential Hebbian learning: Capturing temporal relations between events in neural networks and the brain. Plos Computational Biology, 14, e1006227.

Zhao, Y., Rothkopf, C., Triesch, J. & Shi, B. (2012). A unified model of the joint development of disparity selectivity and vergence control. In Movellan, J. & Schlesinger, M. (eds.), IEEE International Conference on Development and Learning-EpiRob 2012 (ICDL-EpiRob-2012), e1-6. 7-9 November 2012. San Diego, California, USA.

Zlatev, J. & Balkenius, C., (2001). Introduction: Why Epigenetic Robotics? In Balkenius, C., Zlatev, J., Kozima, H., Dautenhahn, K. & Breazeal, C. (eds.), Proceedings of the First International Workshop on Epigenetic Robotics, 1-4, Lund University Cognitive Studies, 85. Lund.